\documentclass[11pt]{article} 
\usepackage{rldmsubmit,palatino}
\usepackage{graphicx}
\usepackage{algorithm}
\usepackage{algorithmic}
\usepackage{amsmath}
\usepackage[utf8]{inputenc} 
\usepackage[T1]{fontenc}    
\usepackage{hyperref}       
\usepackage{url}            
\usepackage{booktabs}       
\usepackage{amsfonts}       
\usepackage{nicefrac}       
\usepackage{microtype}      
\usepackage{xcolor}         
\usepackage{caption}        

\title{CLIP-RL: Aligning Language and Policy Representations for Task Transfer in Reinforcement Learning}

\author{
Chainesh Gautam\\
Department of Data Science and Artificial Intelligence\\
International Institute of Information Technology\\
Bangalore, IN 560100 \\
\texttt{chainesh.gautam@iiitb.ac.in} \\
\And
Raghuram Bharadwaj Diddigi \\
Department of Data Science and Artificial Intelligence \\
International Institute of Information Technology\\
Bangalore, IN 560100 \\
\texttt{raghuram.bharadwaj@iiitb.ac.in} \\
}

\begin{document}

\maketitle

\begin{abstract}

Deep Reinforcement Learning (RL) provides algorithms for solving complex sequential decision-making problems by leveraging the function approximation capabilities of neural networks. 
Recently, there has been an increasing need to develop agents capable of solving multiple tasks within the same environment, especially when these tasks are naturally associated with language.
For instance, in a warehouse setting, a robot may be tasked with executing instructions like, ``Go to location A, pick up object B, and drop it at location C.'' In such scenarios, it would be beneficial to transfer knowledge from previously trained similar tasks to new ones, rather than training models from scratch for each new task.  A straightforward approach in this setting might involve extracting sentence embeddings of natural language instructions, identifying the closest pre-trained instruction using similarity metrics (such as cosine similarity), and initializing the new task's neural network with the trained weights of the closest pre-trained task. However, this method is limited by the assumption that language similarity implies policy similarity, which is not always valid.
In this work, we propose a novel approach that leverages combinations of pre-trained (language, policy) pairs to establish an efficient transfer pipeline. Our algorithm is inspired by the principles of Contrastive Language-Image Pretraining (CLIP) in Computer Vision, which aligns representations across different modalities under the philosophy that ``two modalities representing the same concept should have similar representations.'' The central idea here is that the instruction and corresponding policy of a task represent the same concept, the task itself, in two different modalities. Therefore, by extending the idea of CLIP to RL, our method creates a unified representation space for natural language and policy embeddings. Experimental results demonstrate the utility of our algorithm in achieving faster transfer across tasks.
\end{abstract}

\keywords{
Reinforcement Learning, Transfer Learning, Contrastive Training
}


\startmain 






\section{Introduction}
We consider a setting where an autonomous agent is tasked with learning a set of tasks using a deep reinforcement learning (RL) algorithm. Each task is succinctly described by a natural language instruction, providing a high-level specification of the goal. A key question in this context is: ``When training on a new task, can we leverage knowledge from a subset of tasks the agent has already been trained on?'' Efficient transfer is critical in such scenarios, as it can significantly reduce training time and resources. One popular approach to enable transfer is to initialize the neural network for the new task intelligently. A simple method is to identify the pre-trained task whose natural language instruction is most similar to the new task’s instruction (using a similarity metric like cosine similarity) and initialize the network with the corresponding learned weights. However, this approach inherently assumes that natural language similarity directly correlates with policy similarity, an assumption that is not necessarily true in practice. 

\begin{minipage}{0.684 \textwidth}
For example, consider a simple grid-world environment illustrated in Figure \ref{fig:gridworld}. This environment consists of six tasks where the agent’s objective is to navigate to specific locations based on objects identified by a combination of color and shape. Assume the agent has been trained on five tasks, and the sixth task is “go to the red cone.” The natural language embeddings for this new instruction might be similar to those for instructions like “go to the red box” or “go to the blue cone” based on standard embedding similarity metrics. However, the structure of the environment organizes objects by color rather than shape, making it more appropriate to prioritize policy similarity derived from color-based navigation. Therefore, the challenge lies in tuning instruction embeddings to reflect the structural properties of the environment, ensuring that transfer aligns with task-specific dynamics rather than superficial language similarity. Notice that the weights of the neural networks corresponding to tasks grouped by color, rather than shape, are more likely to exhibit similar patterns. This is because tasks that share structural properties within the environment, such as navigating based on color, lead to similar policies. 
\end{minipage}
\hfill
\begin{minipage}{0.27\textwidth}
    \centering
    \includegraphics[width=\linewidth]{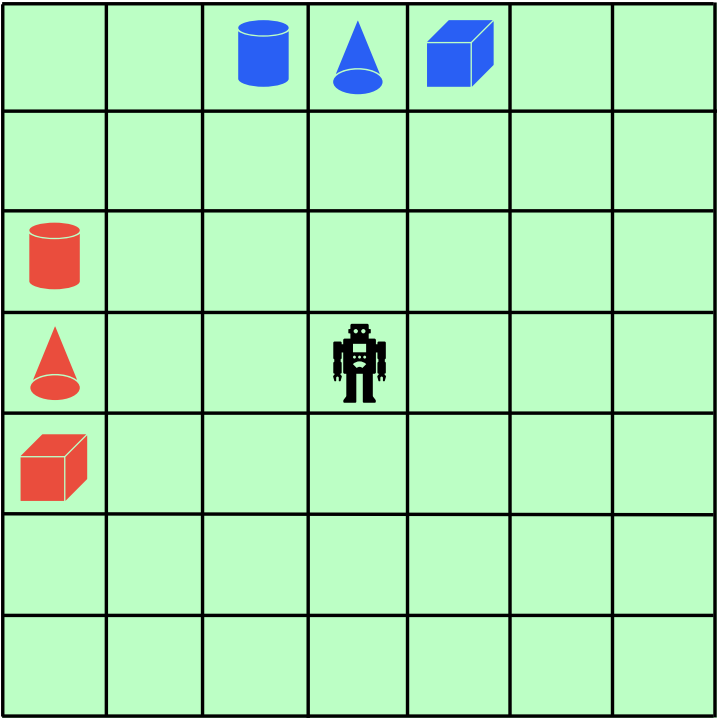}
    \captionof{figure}{An illustration of grid-world}
    \label{fig:gridworld}
\end{minipage}
Therefore, an intelligent transfer approach should leverage this inherent similarity when tuning instruction embeddings. Specifically, the textual description of a task and the neural network weights representing the corresponding policy embody the same underlying concept but in different modalities: language and policy representations, respectively.
This dual-modality representation naturally lends itself to a Contrastive Language-Image Pretraining (CLIP) \cite{radford2021learningtransferablevisualmodels} inspired framework. In CLIP, text and image embeddings are aligned so that pairs referring to the same concept are closer in representation space. Analogously, we propose aligning natural language embeddings of task instructions with the weights of the corresponding policy networks. By applying a contrastive learning objective, instruction embeddings are tuned to reflect task-specific policy similarities rather than superficial linguistic resemblance.

As a result, after tuning, the embeddings of instructions capture the structure of the environment, ensuring that tasks with similar policies are appropriately aligned, even if their language descriptions are not closely related. This alignment significantly improves transfer performance, reducing reliance on language-based similarity and enabling more effective task generalization.

To demonstrate the effectiveness of our approach, we conduct experiments on grid-world environments of varying and increasing sizes. Our results show that the CLIP-inspired transfer model reduces training time by approximately $50\%$ on average compared to a baseline model that relies solely on natural language similarity for task transfer. Additionally, we observe that as the grid size increases, the performance gains from our method become even more pronounced. This highlights the scalability and robustness of our approach, particularly in more complex environments where naive language-based transfer methods are less reliable.
\section{Related Works}
Since the time immemorial, humans make use of instruction in more natural form, namely, language to solve difficult problems using the knowledge of relatively similar problem. Meanwhile, teaching agents how to solve a new task is a central problem in Reinforcement Learning. One of the approach to solve this problem is using natural language instruction to design rewards 
\cite{goyal2019usingnaturallanguagereward, baumli2024visionlanguagemodelssourcerewards}
, but as the complexity of the tasks grow it's challenging to give intricate details of the task using natural language. Another paradigm, called imitation learning \cite{ARGALL2009469}, requires demonstration(s) of the similar desired task, which can then be used to learn the policy for the desired task. However, for each new task, getting the set of demonstration requires lot of human efforts and is not scalable.

In the realm of grounding natural language instructions for robotic systems, recent efforts have focused on mapping human commands to actionable representations that facilitate efficient task planning and execution.
\cite{Arumugam_2017} interprets and executes natural language commands at varying levels of abstraction within a hierarchical planning framework to improve grounding accuracy and \cite{8460937} uses sentences to learn compositional reward function using lambda calculus. It's difficult to scale to complex instructions and environments because these approaches use predefined objects, spatial relations and language based features. \cite{bahdanau2019learningunderstandgoalspecifications} uses an adversarial learning framework where a discriminator differentiates between a predefined set of good (instruction, state) pairs and those generated by the current policy. The output of the discriminator is being used as a reward function to jointly optimize and improve the policy. The main difference between this approach and our approach is, it jointly trains the RL algorithms with linguistic features to improve policy, whereas in our method language is not a part of RL algorithm, rather (language, policy) pairs are being used to initialize policy for similar task. 

Recent works \cite{ fan2022minedojo,goyal2019usingnaturallanguagereward,baumli2024visionlanguagemodelssourcerewards} have introduced intermediate rewards as a means to guide reinforcement learning agents more effectively.
However training these algorithm requires significant human intervention for generating expert trajectories. 
The main difference in methods proposed above and our method is using the language to directly train policies rather than explicitly learning reward function. Our work is closest to  \cite{hutsebautbuysse2019fasttaskadaptationtaskslabeled}, where a set of base control policies along with task description in natural language can be used for task adaptation. However,  \cite{hutsebautbuysse2019fasttaskadaptationtaskslabeled} uses binary classification to sample task description for task adaption and our work makes use of CLIP \cite{radford2021learningtransferablevisualmodels} which is proven effective in contrastive learning between two different modalities and thus helps in selecting similar task description combined with policies.
\section{Problem Description}
\label{gen_inst}

Our objective is to efficiently initialize the policy network for a target task, guided by a natural language description, to accelerate subsequent training. To achieve this, we leverage the language instructions and policy networks from previously trained tasks.

We operate in a Markov Decision Process (MDP) setting where an agent sees the environment via set of states $s \in S$, acts on it with actions $a \in A$, observes the next states with transition probabilities $s_{t+1} \sim P(s_{t+1} | s_t , a_t) $, and receives a reward $r_t \sim R$ with $\gamma \in (0,1]$ as discount factor. In addition to the standard MDP, our agent is provided with a natural language description of the tasks. Let $z$ represent the complete set of tasks to be learned within the environment.

\begin{algorithm}
\caption{}\label{alg:cap1}
\begin{algorithmic}
\STATE $\alpha$ : sample $l$ descriptions from a set of z task descriptions
\STATE $\beta$ : sample $l'$ descriptions from a set of z task descriptions
\STATE $\tau(l)$ : takes sentence as input and gives corresponding vector embedding as output 
\STATE $nn(\pi)$: neural network weights corresponding to the policy $\pi$
\hfill
\STATE Step: 1 (Train Base Policies)
\FORALL{instruction $i \in \alpha$}
\STATE Train base policy $\pi_{i}$ until convergence
\ENDFOR 
\hfill
\STATE Step: 2 (Identify similar language descriptions for transfer)
\FORALL{instruction $j \in \beta$}
\FORALL{instruction $i \in \alpha$}
    \STATE $d_j^i \leftarrow \cos(\tau(j), \tau(i)) = \frac{\tau(j) \cdot \tau(i)}{\|\tau(j)\| \|\tau(i)\|}$
\ENDFOR
    \STATE $ nn(\pi') \leftarrow \frac{\sum_{i} d_j^i nn(\pi_{i})}{\sum_{i} d_j^i}$ 
\ENDFOR
\STATE Initialize the policy network of target task as $nn(\pi')$.

\end{algorithmic}

\end{algorithm}

\section{Proposed Algorithms}
\label{headings}

We begin by discussing the algorithm that employs transfer solely on language-based similarity. Subsequently, we introduce our novel CLIP-inspired transfer technique, designed to address the limitations of the baseline approach.


\subsection{Task adaptation using language similarity}
The fundamental approach of this algorithm is to find similar language semantics between language description of given target task and each source (pre-trained) task. To accomplish this, we use cosine similarity as measure of similarity. Later, we use this similarity to initialize the policy parameters of target task by weighted average of source task policy parameters, weighted by normalized similarity scores. The pseudo-code of discussed algorithm is given in Algorithm \ref{alg:cap1}.

Algorithm \ref{alg:cap1} captures the similar language semantics based on cosine similarity, but doesn't capture its association with policy. As a result, it does not capture the structure of the environment. In the next section, we describe a solution to mitigate this problem. 


\subsection{Task adaptation using CLIP }
The central idea is that the natural language instruction and the optimal policy for a given task represent the same underlying concept, where the policy serves as the actionable realization of the task. Therefore, the embeddings of the language description and the policy representation should be aligned and closer in the shared representation space. 
Initially, we create the dataset $N \ (l_{i},\pi_{i})$ by pairing instructions with corresponding trained optimal base policies. Subsequently, we train CLIP \cite{radford2021learningtransferablevisualmodels} on $N \times N \ (l_{i} \times \pi_{j})_{i,j=1}^{N}$ pairs. CLIP learns a multi-modal shared embedding space by training the policy encoder and instruction encoder, maximizing cosine similarity of policy and instruction embedding of the N diagonal pairs, simultaneously minimizing the cosine similarity of $N^2 - N$ non-diagonal pairs. Given the instruction for target task, we project the instruction 
embedding into shared embedding space and check the CLIP similarity with each instruction in pre-trained dataset. Subsequently, corresponding policies are weighted by normalized CLIP similarity score. We initialize the policy parameters of a given target task by summation of these weighted polices.   
The pseudo-code of our approach is given in Algorithm \ref{alg:cap2}. 
\begin{algorithm}
\caption{}\label{alg:cap2}
\begin{algorithmic}
\STATE $\alpha$ : sample $l$ descriptions from a set of z descriptions
\STATE $\beta$ : sample $l'$ descriptions from a set of z descriptions
\STATE $\tau(l)$ : takes sentence as input and gives vector embedding representation as output 
\STATE $nn(\pi)$: neural network weights corresponding to the policy $\pi$
\STATE $Proj(\tau(l)), ~Proj(nn(\pi))$: Projection of the language embeddings and policy weights, respectively, randomly initialized
\hfill
\STATE Step: 1 (Train $N$ base policies)
\FORALL{instruction $i \in \alpha$}
    \STATE Train base policy $\pi_{i}$ until convergence
\ENDFOR
\hfill
\STATE Step 2: (Prepare dataset for CLIP training and calculate loss) \\
\hspace*{2em} a. Compute the similarity matrix \\
\hspace*{4em} 
$$
S_{ij} = \langle Proj(\tau(l_i)), ~Proj(nn(\pi_j) \rangle \quad \forall i, j \in \{1,2, \dots, N\} ~ (\text{where} ~\langle.\rangle ~ \text{is dot product)}
$$ 

\hspace*{2em} b. Train Projections on the following loss function ($\delta$ is the temperature parameter)
\hspace*{4em} 
$$
L = -\frac{1}{2N} \sum_{i=1}^N \log \frac{\exp\left(S_{ii} / \delta\right)}{\sum_{j=1}^N \exp\left(S_{ij} / \delta\right)}
   -\frac{1}{2N} \sum_{j=1}^N \log \frac{\exp\left(S_{jj} / \delta\right)}{\sum_{i=1}^N \exp\left(S_{ij} / \delta\right)}
$$
\STATE Step: 3 (Identify similar language descriptions for transfer)
\FORALL{instruction $j \in \beta$}
    \FORALL{instruction $i \in \alpha$}
        \STATE $d_j^i \leftarrow \cos(Proj(\tau(j), Proj(\tau(i)) = \frac{Proj(\tau(j) \cdot Proj(\tau(i)}{\|Proj(\tau(j)\| \|Proj(\tau(i)\|}$

    \ENDFOR
       \STATE $ nn(\pi') \leftarrow \frac{\sum_{i} d_j^i nn(\pi_{i})}{\sum_{i} d_j^i}$ 
\ENDFOR
\STATE Initialize the policy network of target task as $nn(\pi')$.
\end{algorithmic}
\end{algorithm}

The main focus of Algorithm \ref{alg:cap2} is to bring both policy and it's corresponding natural language instruction into similar embedding space, so that the agent not only follows the policy which are closer to similar tasks but also understands the language semantics describing it. Algorithm \ref{alg:cap2} not only performs above par the Algorithm \ref{alg:cap1} in terms of computation time but also captures the policies of source tasks which are in the neighbourhood of given target task. 



\section{Experiments and Results}
\label{heading}
We consider grid environments of varying sizes: $8 \times 8$, $10 \times 10$, $15 \times 15$, and $25 \times 25$ for our experimental evaluation. In each environment, the agent is placed at a random position on the grid. The task for the agent is to reach the Goal in the fewest possible steps.  
\begin{figure}
    \centering
    \includegraphics[width=.6\linewidth]{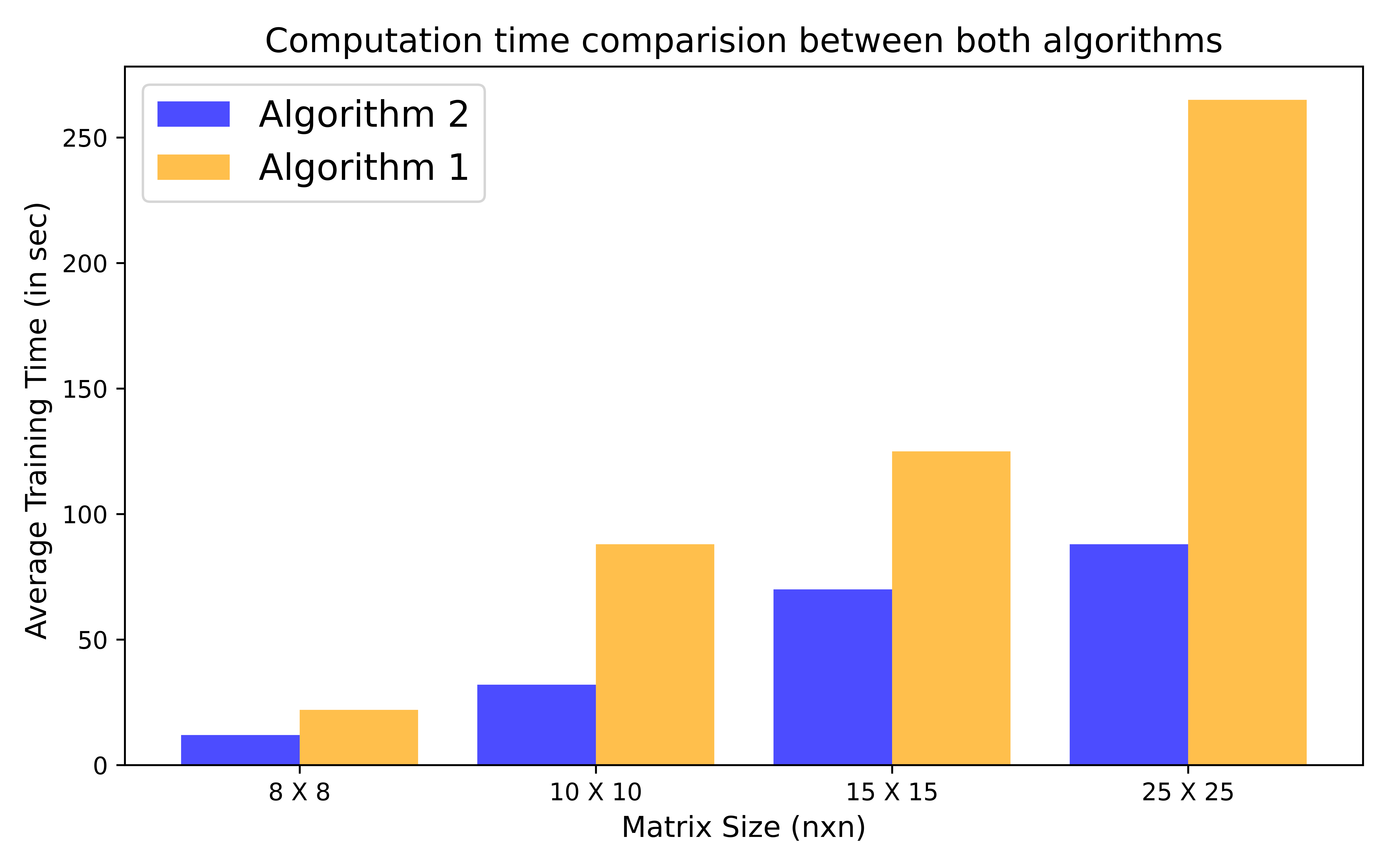}
    \caption{Transfer Performance comparison on grid world of varying sizes. We can observe that CLIP inspired Algorithm \ref{alg:cap2} outperforms Algorithm \ref{alg:cap1}, which is simple language-based transfer.}
    \label{fig:result}
\end{figure}
For all grid configurations and both algorithms, we use four natural language instructions as base tasks: ``top left first'', ``top left second'', ``top right first'', and ``top right second''. Each base task is trained optimally. The target task for transfer is ``top left third''. These instructions represent specific goal positions in the grid. For instance, in a $10 \times 10$ grid, the instruction ``top right first'' places the goal at $(0, 9)$, while ``top right second'' places it at $(0, 8)$. For the target task, the goal is positioned at $(0, 7)$.

We run both algorithms for $10$ independent trials and report the average transfer performance in Figure \ref{fig:result}. It can be observed that Algorithm \ref{alg:cap2} (inspired by CLIP) selects better policies, providing more effective initializations and thus enabling faster training of the target task compared to Algorithm \ref{alg:cap1}. Moreover, as the size of the grid increases, the performance improves exponentially, demonstrating the scalability of the Algorithm \ref{alg:cap2}.

\bibliographystyle{plain}
\bibliography{refer.bib}

@misc{radford2021learningtransferablevisualmodels,
      title={Learning Transferable Visual Models From Natural Language Supervision}, 
      author={Alec Radford and Jong Wook Kim and Chris Hallacy and Aditya Ramesh and Gabriel Goh and Sandhini Agarwal and Girish Sastry and Amanda Askell and Pamela Mishkin and Jack Clark and Gretchen Krueger and Ilya Sutskever},
      year={2021},
      eprint={2103.00020},
      archivePrefix={arXiv},
      primaryClass={cs.CV},
      url={https://arxiv.org/abs/2103.00020}, 
}

@misc{bahdanau2019learningunderstandgoalspecifications,
      title={Learning to Understand Goal Specifications by Modelling Reward}, 
      author={Dzmitry Bahdanau and Felix Hill and Jan Leike and Edward Hughes and Arian Hosseini and Pushmeet Kohli and Edward Grefenstette},
      year={2019},
      eprint={1806.01946},
      archivePrefix={arXiv},
      primaryClass={cs.AI},
      url={https://arxiv.org/abs/1806.01946}, 
}

@misc{goyal2019usingnaturallanguagereward,
      title={Using Natural Language for Reward Shaping in Reinforcement Learning}, 
      author={Prasoon Goyal and Scott Niekum and Raymond J. Mooney},
      year={2019},
      eprint={1903.02020},
      archivePrefix={arXiv},
      primaryClass={cs.LG},
      url={https://arxiv.org/abs/1903.02020}, 
}

@article{ARGALL2009469,
title = {A survey of robot learning from demonstration},
journal = {Robotics and Autonomous Systems},
volume = {57},
number = {5},
pages = {469-483},
year = {2009},
issn = {0921-8890},
doi = {https://doi.org/10.1016/j.robot.2008.10.024},
url = {https://www.sciencedirect.com/science/article/pii/S0921889008001772},
author = {Brenna D. Argall and Sonia Chernova and Manuela Veloso and Brett Browning}
}

@misc{baumli2024visionlanguagemodelssourcerewards,
      title={Vision-Language Models as a Source of Rewards}, 
      author={Kate Baumli and Satinder Baveja and Feryal Behbahani and Harris Chan and Gheorghe Comanici and Sebastian Flennerhag and Maxime Gazeau and Kristian Holsheimer and Dan Horgan and Michael Laskin and Clare Lyle and Hussain Masoom and Kay McKinney and Volodymyr Mnih and Alexander Neitz and Dmitry Nikulin and Fabio Pardo and Jack Parker-Holder and John Quan and Tim Rocktäschel and Himanshu Sahni and Tom Schaul and Yannick Schroecker and Stephen Spencer and Richie Steigerwald and Luyu Wang and Lei Zhang},
      year={2024},
      eprint={2312.09187},
      archivePrefix={arXiv},
      primaryClass={cs.LG},
      url={https://arxiv.org/abs/2312.09187}, 
}

@inproceedings{Arumugam_2017, series={RSS2017},
   title={Accurately and Efficiently Interpreting Human-Robot Instructions of Varying Granularities},
   url={http://dx.doi.org/10.15607/RSS.2017.XIII.056},
   DOI={10.15607/rss.2017.xiii.056},
   booktitle={Robotics: Science and Systems XIII},
   publisher={Robotics: Science and Systems Foundation},
   author={Arumugam, Dilip and Karamcheti, Siddharth and Gopalan, Nakul and Wong, Lawson and Tellex, Stefanie},
   year={2017},
   month=jul, collection={RSS2017} }

@INPROCEEDINGS{8460937,
  author={Williams, Edward C. and Gopalan, Nakul and Rhee, Mine and Tellex, Stefanie},
  booktitle={2018 IEEE International Conference on Robotics and Automation (ICRA)}, 
  title={Learning to Parse Natural Language to Grounded Reward Functions with Weak Supervision}, 
  year={2018},
  volume={},
  number={},
  pages={4430-4436},
  doi={10.1109/ICRA.2018.8460937}}

@article{article,
author = {Fan, Linxi and Wang, Guanzhi and Jiang, Yunfan and Mandlekar, Ajay and Yang, Yuncong and Zhu, Haoyi and Tang, Andrew and Huang, De-An and Zhu, Yuke and Anandkumar, Anima},
year = {2022},
month = {06},
pages = {},
title = {MineDojo: Building Open-Ended Embodied Agents with Internet-Scale Knowledge}
}

@misc{hutsebautbuysse2019fasttaskadaptationtaskslabeled,
      title={Fast Task-Adaptation for Tasks Labeled Using Natural Language in Reinforcement Learning}, 
      author={Matthias Hutsebaut-Buysse and Kevin Mets and Steven Latré},
      year={2019},
      eprint={1910.04040},
      archivePrefix={arXiv},
      primaryClass={cs.AI},
      url={https://arxiv.org/abs/1910.04040}, 
}

@inproceedings{fan2022minedojo,

  title = {MineDojo: Building Open-Ended Embodied Agents with Internet-Scale Knowledge},

  author = {Linxi Fan and Guanzhi Wang and Yunfan Jiang and Ajay Mandlekar and Yuncong Yang and Haoyi Zhu and Andrew Tang and De-An Huang and Yuke Zhu and Anima Anandkumar},

  booktitle = {Thirty-sixth Conference on Neural Information Processing Systems Datasets and Benchmarks Track},

  year = {2022},

  url = {https://openreview.net/forum?id=rc8o_j8I8PX}

}
\end{document}